# The presence of occupational structure in online texts based on word embedding NLP models

Zoltán Kmetty[1,2]

Júlia Koltai[1,3]

Tamás Rudas[2]

1. Centre for Social Sciences – MTA Centre of Excellence, CSS-Recens Research Group. 1097, Budapest, Tóth Kálmán u. 6, Hungary

2. Eötvös Loránd University, Faculty of Social Sciences. 1117, Budapest, Pázmány Péter sétány 1/A, Hungary

3. Central European University. Quellenstraße 51, 1100 Wien, Ausztria

# Abstract

Research on social stratification is closely linked to analysing the prestige associated with different occupations. This research focuses on the positions of occupations in the semantic space represented by large amounts of textual data. The results are compared to standard results in social stratification to see whether the classical results are reproduced and if additional insights can be gained into the social positions of occupations. The paper gives an affirmative answer to both questions.

The results show fundamental similarity of the occupational structure obtained from text analysis to the structure described by prestige and social distance scales. While our research reinforces many theories and empirical findings of the traditional body of literature on social stratification and, in particular, occupational hierarchy, it pointed to the importance of a factor not discussed in the main line of stratification literature so far: the power and organizational aspect.

# Introduction

Analysis and positioning of occupations in social space has a long history in social research. Social stratification models use occupations as a standard way of operationalizing the position of people in the society. Most of the stratification models rely on massive survey data. However, the developments of information technology, in particular data science and natural language processing (NLP), and also the rapid growth of computing capacity provide new types of data sources. NLP methods – like word embedding used in this analysis – open up the opportunity to examine the society through written/digitalized texts.

The language used by a social group, mirrors the group's cultural frame of mind (Kozlowski et al. 2019). These texts inform us about the ways of thinking, feeling and knowledge of people. (Evans-Aceves 2016). Billions of digitalized or originally digital texts are available for analysis, which all depict mentality, opinion and values. Sources of texts vary from social media posts, through online newspapers and forums to whole books of classic literature or scientific papers. Thus, the analysis of these huge corpuses can help the understanding of people's perceptions and ways of thinking in a given culture about any kind of topic.

Our paper focuses on the positions of occupations in the semantic space represented by large amounts of textual data. The results are compared to standard results in social stratification to see whether the classical results are reproduced and if additional insights can be gained into the social positions of occupations. The paper gives an affirmative answer to both questions.

The main contribution of this paper is that social structures, in particular, stratification of occupations – established so far based on purposively collected data –, do exist and can be derived from large text corpora using methods of unsupervised learning. Further, the most important factors organizing this stratification can be implied, not from theoretical considerations, rather from the semantic space depicted in the text corpora.

In the first part of the paper we briefly introduce a review how social scientists measure the position of people in the society. We also discuss the basics of NLP and especially word embedding models and give a short review on how occupations have been analysed so far using NLP methods. In the Data and Methods chapter, we describe the large digitalized corpora we have used in the analysis and the specification of the model, with which we have extracted the latent dimensions of occupations from these corpora. This part is followed by the analysis and the results. The paper closes with a discussion of how these findings reinforce and extend our understanding of the societal positions of occupations.

**Theoretical Background**

*Occupations and social structure*

Social class and social stratification are widely used concepts from the early years of sociology. Some variants of these concepts are theory driven, others rely on empirical data. Some of them use categories to describe people's positions in the social structure, others apply continuous scales. In social stratification research, occupation is routinely used to link the positions of the individuals to their memberships in a social stratum. In industrialized societies, occupation is a very strong indicator of social standing and as it tends to be more stable than income, it serves as a much better proxy for the position of an individual (Connelly et al 2016). Thus, the goal of these researches is to classify the occupations in a way, which mirrors the stratification of the society. There are multiple approaches exist regarding the measurement of occupational position in social space. Some theories use occupation for the creation of vertical hierarchies with continuous scales (Ganzeboom-Treiman 1996); others use it for the creation of discrete stratification categories with both horizontal and vertical dimensions (Goldthorpe et al 1982, Rose-Harrison 2007). Researchers used various measurements for the classification of occupations to create their stratification models. Based on Bukodi et al (2011), we can divide these measurements to two types: one type of the measurements uses subjective indicators, the other type works with objective indicators. The scale of Goldthorpe and Hope (1972) belongs to the former category. They applied a synthetic scale of subjective opinions to measure the general desirability of occupations. Treiman (1977) also used questions on subjective perceptions and from these he created the Standard International Occupational Prestige Scale (SIOPS), which is a widely used analytical scale. The International Socio-Economic Index (ISEI) (Ganzeboom-Treiman 1996) and the Cambridge scale (Prandy-Lambert 2013, and Meraviglia et. al 2016) are good examples for the other type of the scales, which use objective data in their measurement. ISEI builds on educational level and mean income of the occupations to create their hierarchy. The Cambridge scale uses the marriage-table based social distance of occupations to map their hierarchy. Chan and Goldthorpe (2004) applied similar methodology, but their one was built on close friendship data, and not marriage tables. In their interpretation the scale measures the hierarchy of social status. Meraviglia and her colleagues (2016) argue that all continuous measure of social stratification are the indicators of the same latent dimensions.

But which characteristics of the occupations matter? The answer varies from one social stratification model to the other. The Erikson-Goldthorpe-Portocarero (Erikson et al 1979) (EGP) model – which is one of the well-known occupation-based stratification models – is built on the on the employment relations in labour market. The market and the work situation (e.g. level of income, economic security, authority level) are the dimensions, which determine the class position. (Connelly et al 2016) Along with education, income also plays an important role in the construction of the ISEI scale. In the case of the SIOPS scale, occupations are ordered by their prestige, which is measured by the subjective judgement of respondents of large-scale surveys.

In this paper, we explore how occupation structure could be measured through online texts. Our approach is a data-driven one, as we unfold the different layers of occupational structure in online digitalized texts, and not on purposively collected data. From this viewpoint, the closest model from the abovementioned ones is the Cambridge Scale. However, we do not focus on the social ties, but rather on the semantic ties of the occupations. In the next subchapter, we introduce those novel text mining techniques, with which we can examine the semantic ties of the occupations and through these, study the structure of the society.

*Text as data and word embedding models*

Process produced data, like text messages, phone calls, the usage of public transport with digital tickets, social media posts, bank transfers all leave digital marks in databases of different systems. These data are not generated by the users with the understanding that they will be part of some analyses, thus, these data mirror the behaviour of individuals better than data from classical surveys or other research, where self-reported responses can be biased by the interview situation, by social desirability and by limitations to recall past events. (Lazer and Radford 2017) For that very reason, the analyses of these data can be exceptionally interesting for social research.

This information is stored in very diverse formats, from pictures, through videos or voices, to numbers and the majority of these data are stored or can be transformed into textual formats. Text analysis has always had an important place in the field of sociology. From the line-by-line reading and analysis at the birth of the science, through coding and linking the text by the researcher (Bales 1950) to digital and partly automatized coding of smaller corpora (Hays 1960), it was always part of sociology – which, according to Savage and Burrows (2007), defined its expertise through its own methods. However, these classic analytical methods could not handle large-scale corpora with thousands of millions of words. The methodological knowledge needed for the analysis of large text data had to be imported from computational linguistics, data- and computer science. Parallel with the increase of in the amount of digital data, computational power and artificial intelligence have also developed. New methods, which aim the processing of large digital corpora, emerge and are continuously elaborated. These methods have to be incorporated by sociologists, otherwise, they would miss the opportunity of interpreting such sources of data.

Just like partly automatized methods of earlier times, automated text analysis and natural language processing combine qualitative and quantitative approaches. The latest methods provide the deepness of qualitative analysis with the advantage of large number of observations in quantitative analysis. However, one of the consequences, that these textual data mostly record observed behaviour, is that its structure and relevance (its 'noisiness') is not as appropriate for analyses as of data collected by traditional techniques. The phase of data cleaning and structuring includes important decisions of the researcher. These decisions can influence the inner and outer validity of the results, thus the very detailed documentation and the description of the arguments behind these decisions are extremely important for making the research transparent.

Simpler methods of text analysis only focus on the words of the corpus, as if they had no relations with the surrounding words and sentences, but more complex methods can also take the structure of the text into account. Some of these methods are based on the 'bag-of-words' model, which means that words are treated together with their environments, namely a given number of words around them. The size (the number of words) of the environment is defined by the researcher and can be any positive integer, though too wide environments can cause loss of context. The examination of the environment is proceeded for each word, as a sliding window through the whole corpus, and the result of the method is based on the complex co-occurrences of words.

The method we used in our analysis is a neural network-based word embedding model (Mikolov et al 2013). This method helps the researcher to understand the deeper meaning of texts with modelling the semantic meaning of words. The position of a word is defined by its context, which approach has non-computerized linguistic theoretical base, originated by Firth (1957). The word embedding model projects the position of each word of a corpus to a low dimension vector space. The most popular method Word2Vec (Mikolov et al 2013) uses a neural network based logistic classifier to estimate the word positions. The words of the corpus are positioned in this semantical vector space, where we can calculate the contextual proximity of words. This proximity does not only reply on the co-occurrences of the words, but also on the co-occurrences of the contexts of words. Several word embedding methods are available (e.g. Word2Vec by Mikolov et al (2013), Glove by Pennington et al (2014), and Fasttext by Joulin et al (2016)) to train textual data, and to establish proximities[1]. In either method, the positions a word defines its meaning in the semantic space. Two words with similar environments will be close to each other, thus, words with similar meaning will be nearby.

---

[1] The latest generation of word embedding models (like BERT – Devlin et. al 2018) creates contextualized vector spaces and not static ones. They calculate unique word distances for different contexts. These models perform very well in classification tasks, but they are not applicable to analyzes such as ours, which do not look for varying positions of a word in different contexts but targets the general position of words.

Proximities of words are frequently defined by the cosine of the angle formed by the vector of the words. Standard metrics like Euclidean distance could be misleading here, because the length of each word vector is strongly correlating with the frequency of the word within the corpora (and it also depends on the context variability) (Schakel – Wilson 2015).

Kozlowski et al (2019) showed, that these proximities can be successfully used for the analyses of culture. The starting point of their analyses was based on the theoretical foundation, that language (and texts) mirrors the way of thinking of those, who uses them. Thus, the analysis of written texts makes researchers able to draw conclusions about the society the texts originate from. They showed that in word embedding methods, we can create dimensions of social inequality with the proximity of words, which represent the two extreme values of a given inequality (e.g., poor – rich; male – female). Mirroring this proximity to other words, we can unfold hidden inequalities of the society. For example, if we mirror the dimension of gender to the word 'doctor', we find the word 'nurse' at the other end of the dimension. The gender inequality of these medical professions can be captured in the vector space. (See similar results of Bolukbasi et al 2016, Caliskan et al 2017, and Garg et al 2018) For sociologist, these analogies can help a lot in the understanding of social cleavages, as based on the concept of Kozlowski et al. (2019), they can unfold unconscious or not yet proved patterns of social inequalities.

## Data and Methods

In the previous section we presented the basics of word embedding models and showed how these models can be used to analyze social phenomena. In this research we use pre-trained word vector models. These widely used word vectors are publicly available, which makes our results reproducible. The embeddings are trained on large scale corpuses, which is important as previous research showed that the accuracy and validity of word embedding (measured on word analogies) highly depend on the corpus size (Mikolov 2013). These pre-trained vector spaces are frequently used in NLP tasks. But previous studies have also confirmed that these vector space models can be used well to study social processes and social context as well. Researchers have validated with surveys, that vector space models trained on large and general corpus can be used to measure cultural patterns (Kozlowski et al 2019) or even stereotypes against social groups (Joseph – Morgan 2020).

We used three pre-trained vector spaces in the analysis. The first vector model we used was trained on the English language texts of the Common Crawl (CC) corpus[i], a huge web archive, which contains raw web page data, metadata and text extractions. The raw web pages can be everything, from a news site, through a blog or a page of a university, to pages like Amazon Books. As the authors state, they provide "a copy of the internet". It consists of one petabyte of data, collected between 2011 and 2017. The word embedding model was trained on the English language pages of this corpus. As the data do not contain geo-location of the websites, they might include web sites from all over the world. In the initial corpus 600 billion tokens were identified and the vector space consists of 2 million words positioned in a 300-dimensional space[ii]. The training of the corpus was realized by Fasttext algorithm (Joulin et al 2016)

The second vector space we used is the Wikinews, which was trained on a combined corpus of the English Wikipedia (saved in 2017), the UMBC WebBase corpus, and another corpus, which contains all the news from stamt.org. The UMBC corpus contains high quality English paragraphs derived from the Stanford WebBase project and contains 100 million web pages from 2007. Statmt.org contains political and economic commentary crawled from the web site Project Syndicate. The combined corpus is quite diverse and has 16 billion tokens. The vector space consists of one million words, positioned in a 300-dimensional space and was trained by Fasttext algorithm (Joulin et al 2016). Thus, the number of dimensions and the training method of the two vector spaces were the same.

We used a third vector space, which was also built on combined corpus of the Wikinews sources, but in this third vector space, during the training phase of the model, sub-word information was also taken into account. It means that partly identical or words or words with the same root like sociology and society tend to be closer to each other in this vector space. We will refer to this vector space later as Wikinews Subwords. On this vector space, we utilize the the innovation of the fasttext algorithm, namely that it can account for sub-word information. Although the first two vector-spaces were also trained by fasttext algorithm, sub-word information were not taken into account there: thus, the method was closer to a word2vec solution (which cannot handle subword information).

The pre-trained vectors we used in this paper are trained on general English corpora. We could not narrow the geographical focus, as we do not know the geographical distribution of the authors of texts. However, based on other results in this topic (see Treimann 1977), there are no significant differences between the prestige scores of different developed countries.

Altogether 234 occupations (see table A1 for the list) were selected for the analysis and we used the most common 200,000 words of each vector space. In the ISCO classification, more than 7000 occupations are listed, but the number of one-world length occupations was around 750. We manually checked all these occupations and selected those more than 200, which were not extra unique or rare (like chieftain). These occupations cover both the vertical and horizontal aspects of occupational space. Although we tried to create a gender-balanced occupational list, male occupations are overrepresented based on our qualitative estimations. Some of the pre-selected occupations were not among the most common 200,000 words, so we had to omit them. At the end, from these 234 occupations, 204 occupations were detected in CC, and 207 in Wikinews (202 occupations were available in both corpora). We located the position of each 204 and 207 occupations in the vector spaces. The same methods were applied for each vector spaces (CC, Wikinews. Wikinews Subwords): the cosine-similarities of each pairs of occupations were computed in the 300-dimensional vector space. These cosine-similarities are the ones, which represent the semantic closeness of the occupations. Table 1 shows a small part of the similarity table, based on the CC corpus.

|  | doctor | cardiologist | sociologist | historian | shopkeeper | barmaid |
|---|---|---|---|---|---|---|
| doctor | 1.00 | 0.61 | 0.29 | 0.25 | 0.34 | 0.25 |
| cardiologist |  | 1.00 | 0.32 | 0.27 | 0.20 | 0.11 |
| sociologist |  |  | 1.00 | 0.62 | 0.31 | 0.25 |
| historian |  |  |  | 1.00 | 0.26 | 0.20 |
| shopkeeper |  |  |  |  | 1.00 | 0.48 |
| barmaid |  |  |  |  |  | 1.00 |

Table 1 Semantic closeness of selected occupations (cosine similarity, CC corpus)

As we only focused on similar concepts, namely occupations, all words have positive cosine similarity. (The theoretical range of cosine similarity can go between -1 and 1.) Over this similarity, we can observe large differences in the values of the table. Not surprisingly, the occupation *doctor* is close to *cardiologist*; *sociologist* is close to *historian* and *shopkeeper* is close to *barmaid*. At the same time, *doctor* is distant from *sociologist*, *historian* and *barmaid*, *shopkeeper* is distant from *cardiologist* and *historian*. We can observe that distinct domain areas of occupations can be identified based on the similarity matrix.

As we have mentioned in the Introduction, one of the main goals of our research was to extract the most important dimensions, which structure the occupations in the semantic field. To fulfil this goal, we applied factor analysis with rotation on the similarity matrix – instead of the often applied correlation matrix – as input. As a robustness test, we repeated our computations with different factor analysis methods and different rotation techniques, and the difference between the results were quite small. The presented results are based on a minres (Minimum Residual) factor analysis technique and varimax rotation (Revelle 2018). The following analyses are based on the factor loadings resulted of this methodology.

Due to the exploratory nature of the research, we have not had strong assumption on the number of factors to be extracted. The decision on the number of factors was based on empirical tests and also on practical considerations. We decided to select more than 1 factor as we wanted to understand the most important dimensions behind the structure of the occupations, and not only the main dimension, At the same time, we decided to select maximum of 5 factors, in order to keep the interpretability. Average residuals for the similarity matrix (RMSR value) and Chi-square based fit indices were used to test the statistical validity of the models, and external measures (like ISEI scale) were applied for the comparison of the results to test criterion validity. Overall, we found that all the 2, 3, 4 and 5-factor solutions are worth to investigate. In the later analysis, we detail the 3-factor solution as it looked the most promising one.

We used different methods for the robustness test of the models. We compared the consistency of the results of different vector spaces with the cross-correlation of the factors generated in the different vector spaces. We also tested the similarity of the context of the same words across different vector spaces. If we have two independently trained vector space, the cosines similarity of the same words in the different vector spaces is around 0, as the position of the words is very unlikely to be the same in the two vector spaces. Thus, to compare the context of the words, first we have to align the two vector spaces. For testing the context similarity of the words across different vector spaces, from the most frequent 200,000 words of each corpus we selected those 153 423 words, which appear in both corpora, and we aligned the Wikinews vector space to the Common Crawl vector space with Procrustes rotation. In the aligned Wikinews vector space, the cosine similarity of occupation pairs remained the same, but we could calculate the similarity of the same words in the two embeddings. In a case of perfect alignment, the cosine similarity would be 1. But in real word examples, the similarity never reaches the celling point (1), as the training process add some random variation and also because the context of the words is different. But higher similarity means a higher context stability across embeddings. This alignment technique has been used in previous papers to measure the context variation of different concept through the time (Hamilton – Leskovec – Jurafsky 2016), but in this paper, we primary use this to measure the stability of occupation contexts across embeddings based on different text corpora.

## Heuristics

Before starting the analysis of occupations in the vector spaces, we present some examples about the context of occupations to show it more intuitively, what these models are based on. As the goal is to measure social structure through the semantic position of occupation the most important question is how social structure is presented in textual data. In the examples below we selected some sentences, which include occupations. Here we use cultural examples to present the social position of occupations. But we could replace the cultural examples with o- any other life domain. We ask the reader of this paper to go through these examples and think about if it is possible to change the occupations between the sentences and what is the likelihood that the changed sentence will occur.

**Example 1.**

*Last night the **SENATOR** went to the theatre*

*This evening the **TYPIST** wanted to go bowling.*

We can assume, that different cultural activities are closer to specific occupations – as occupation strongly correlates with status, power and money. A senator might also play bowling but has higher probability to go to the opera or to the theater, than a typist.

**Example 2.**

*Half of the company's **DATA_SCIENTISTS** graduated from Ivy League schools.*

*The plan of the **WAITRESS** was to attend evening school next year.*

The above described situation is the same in the second example. Usually a waitress does not graduate from an Ivy League school, and data scientists do not attend evening schools.

Above the intuitive understanding of these examples, we tested them on our data. We tested the closeness of occupations to certain activities with the cosine similarity of the words of the occupation and the activity. In the CC vector space, the cosine similarity of the occupation senator with the word theatre is 0.21, the same measure for the typist is 0.12. For bowling, the senator's cosine similarity is 0.05 but the typist's value is 0.16. Thus, the senator is closer to the high-end cultural activity, while the typist is closer to the more popular one. These results strengthen the intuitive assumption, namely that in these contexts, the presence of different occupations has different likelihoods.

At the same time, it is important to emphasize the different logic of word embedding similarity and similarities of occupational hierarchies created by social scientists. Table 2 presents two occupational pairs as examples. The ISEI distance of the two occupations in the same row is 0 in both cases, which means that these occupation pairs have the same prestige positions. However, the cosine similarity of these occupation pairs is different in the first and in the second row, which suggests that distances are different in the case of the first and the second row. The reason of this difference lies in the semantic relation of these pairs. In the first row there are two occupations which are different in many aspects, in spite of having the same prestige, while the two occupations in the second row are about identical. Thus, we don't assume to get the exact same results from the word embedding analysis, as from the different occupational scales, like ISEI.

|  | cosine similarity in the CC corpus | ISEI distance |
|---|---|---|
| anatomist - ornithologist | 0.49 | 0 |
| barman - bartender | 0.81 | 0 |

Table 2 cosine similarity and ISEI distance of occupation pairs (example)

**Results**

*Common Crawl*

First, we present the results from the Common Crawl corpus. From the list of the occupations, the doctor was the most frequent item. Overall, it was the 1496th most frequent word in the list of words contained by the corpus. Driver, writer, cook, judge, editor, lawyer, professor or attorney were also frequent. We can observe a pattern, that those occupations are more frequent in this corpus, which have higher prestige.

As we have mentioned above, first, we calculated the cosine similarity of the 204 occupations, which were in the most frequent 200,000 words of the vector space. Then we used this similarity matrix as an input to extract factors, based on which we detected the main structural dimensions of the occupational semantic space. We tested the model for different number of factors. In the case of the two-factor solution the Root Mean Square Residual (RMSR) was 0.07. The explained variances of the two factors were quite similar. In the case of both dimensions, knowledge is an important factor. Based on the occupations with the highest loading on a given factor, the first dimension is closer to the domain of the media (e.g., commentator, editor), and the second is closer to the domain of science. (See Table A2 in the Appendix for more details and information on the highest loadings.) We calculated the correlation of the factor loadings with the ISEI scale[2]. The correlation was 0.64 in the case of the first dimension, and 0.79 in the case of the second dimension – which are quite high, especially in the second case. These results suggest that both dimensions reflect on the vertical positions of occupations.

In the case of a three-factor solution the RMSR value was 0.06. The importance of the dimensions was not as equal as in the previous model with two factors. The first factor has the largest correlation with the ISEI prestige scores (Pearson r = 0.71). The correlations of the second and third dimensions were moderately high, 0.59 and 0.45 respectively. This means that the corpus contains a strong footprint of the hierarchal social structure.

We have also tested the correlation of the factors of the two- and three-factor models. We found that the correlation of the first factors of the two- and three factor solution was 0.9, and the correlation between the second factors was the same.

Table 3 shows the occupations with the highest and lowest factor loadings on a given factor of the three-factor model. Interpreting the three factors, we found that the first two factors were quite similar, but with some important differences. In the case of the first factor the role of institutional power seems to be more important – the chancellor or the dean are good examples for that. The second factor is structured more on the basis of knowledge and educational level associated with the occupations, while the third factor is built up by the dimensions the power levels and organizational capacities of the occupations.

---

[2] A usual way to create a factor model is to start from a raw data source, calculate the covariance/correlation matrix and then calculate the factor loadings and estimate the factor scores based on these loadings. In this paper, we start from a similarity matrix and calculate the factor loadings. As we do not have raw data here, we could not calculate the factor scores. That has one important implication. Rotated factor scores are statistically independent, but factor scores are not. That is why we have a strong correlation between the extracted factors.

| First factor | | | | Second factor | | | | Third factor | | | |
| --- | --- | --- | --- | --- | --- | --- | --- | --- | --- | --- | --- |
| Highest loadings | | Lowest loadings | | Highest loadings | | Lowest loadings | | Highest loadings | | Lowest loadings | |
| rank order | ISEI | rank order | ISEI | rank order | ISEI | rank order | ISEI | rank order | ISEI | rank order | ISEI |
| chairperson | 71.29 | dressmaker | 23.47 | ecologist | 80.46 | courier | 30.34 | secretary | 44.94 | brazier | 28.52 |
| ecologist | 80.46 | electrician | 36.35 | historian | 83.81 | stewardess | 46.76 | commissioner | 78.76 | animator | 79.74 |
| professor | 85.41 | waiter | 25.04 | biologist | 80.46 | waiter | 25.04 | treasurer | 73.38 | tattooist | 50.15 |
| chancellor | 70.34 | shopkeeper | 35.34 | writer | 72.83 | vendor | 23.53 | mayor | 68.77 | plasterer | 18.02 |
| advocate | 86.72 | roofer | 22.16 | philosopher | 83.81 | driver | 26.85 | chancellor | 70.34 | cleaner | 16.38 |
| dean | 65.01 | maid | 14.21 | geographer | 83.09 | babysitter | 24.98 | prosecutor | 86.72 | acrobat | 37.59 |
| director.general | 71.29 | barman | 25.04 | zoologist | 80.46 | cleaner | 16.38 | dean | 65.01 | potter | 24.43 |
| commentator | 72.83 | barmaid | 25.04 | novelist | 72.83 | housemaid | 16.38 | senator | 68.77 | dancer | 61.82 |
| neurologist | 81.92 | housemaid | 16.38 | sociologist | 83.09 | barmaid | 25.04 | rector | 70.34 | painter | 61.82 |
| historian | 83.81 | barber | 31.08 | physicist | 84.61 | brazier | 28.52 | governor | 68.77 | weaver | 28.95 |
| commissioner | 78.76 | plumber | 29.16 | mathematician | 81.78 | constable | 51.5 | chairperson | 71.29 | bender | 25.78 |
| environmentalist | 80.46 | blacksmith | 25.63 | ornithologist | 80.46 | receptionist | 39.02 | clerk | 43.33 | cook | 24.53 |
| curator | 77.19 | plasterer | 18.02 | poet | 72.83 | waitress | 25.04 | attorney | 86.72 | assembler | 27.91 |
| biologist | 80.46 | carpenter | 26.62 | journalist | 72.83 | clerk | 43.33 | congressman | 68.77 | dishwasher | 16.5 |
| sociologist | 83.09 | bricklayer | 22.57 | botanist | 80.46 | maid | 14.21 | constable | 51.5 | welder | 28.52 |

Table 3 Occupations with highest and lowest loadings, 3-factor solution, CC corpus

For a deeper understanding of the results we further analyzed the first dimension of the three-factor solution. In the rest of the paper, we refer to this dimension as Occupation Semantic Position Scale (OSPS).

Figure 1 shows the scatterplot of the ISEI and the OSPS scales. We calculated for all pairs of occupations, if they are in the same rank order in the two scales. The result of this calculation shows, that in 75 percent of the occupation pairs, the order was the same. Thus, we can assume, that the that proximity of occupations in the online texts strongly correlates with the expected educational level and the average income of the selected occupation, which are the basic dimensions of the ISEI prestige score.

We have to emphasize that word embedding method is an unsupervised one, which means that the researchers do not put external information to the model. According to this, we haven't used the ISEI prestige scores as an input of the model, neither we optimized varimax rotation for that. Thus, these results are only based on the information contained in the online texts.

At the same time, we have found remarkable differences. Some occupations like doctor, dentist, pharmacist or solicitor were positioned quite low in the OSPS, while high on the ISES. The reason of it is that the position of an occupation on the OSPS does not only depend on the prestige of the occupation, but rather affected by the reflection of the domain, which surrounds the occupation. For example, being a dentist is a high prestige job, paired with high educational level and high income, but (1) being sick is not a positive situation (which feelings can be mirrored in the texts) and (2) everybody can be sick, irrespective of their social status: health care professionals provide services to the general public, which means they have links to all levels of the social structure. As health-related occupations are all affected by these circumstances, this can be the reason that they are scored lower.

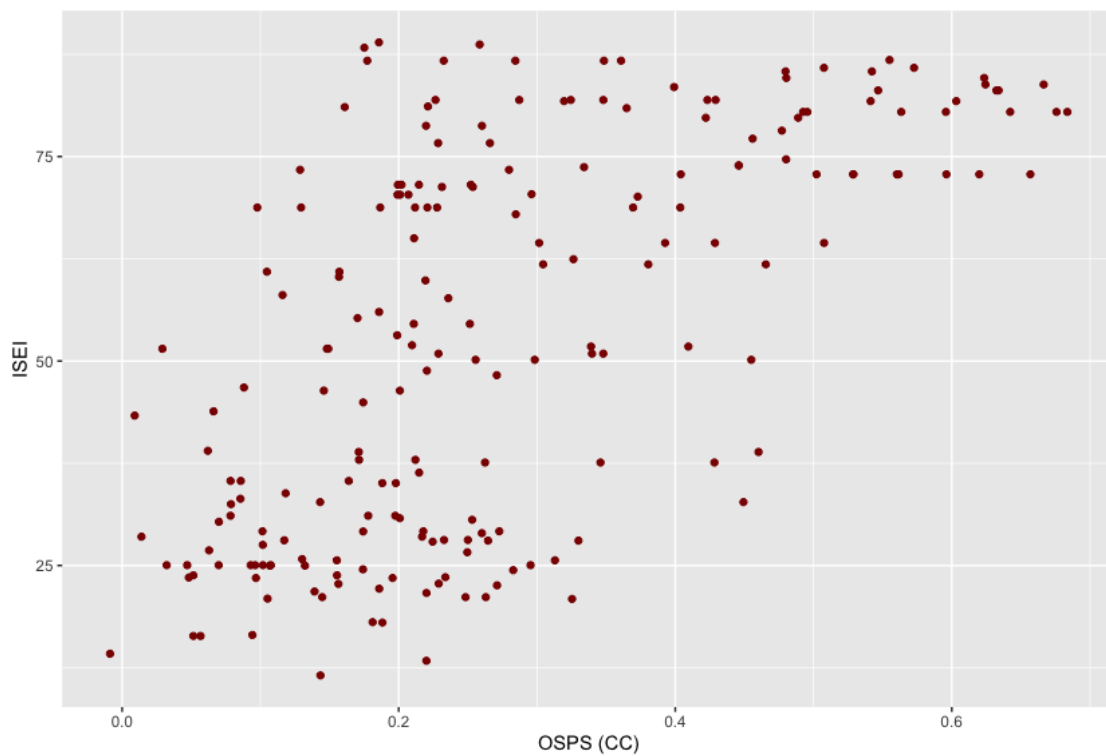

Figure 1 Scatterplot of word embedding based occupation prestige score (Occupation Semantic Position Scale – OSPS) from the CC vector space and ISEI

Knowledge and power level are important factors of prestige, so it is not surprising we found these dimensions behind the hierarchical structure of semantic positions of occupations. At the same time, wage doesn't appear as an organizing principle in this hierarchy, however, it is an important dimension of the prestige scales. As we wanted to know if wage can be detected as a background dimension in further factors, we additionally created 4- and 5-factor models.

In the case of the 4-factor model, the first three dimensions were quite similar to those that we have found in the 3-factor solution. The main structuring dimension of the fourth factor was gender: occupations with the five highest loadings were receptionist, waitress, babysitter, manicurist and hairdresser. In the 5-factor model, we still haven't detected wage as an organizing dimension of any factors. What we have found was that health-related occupations score high on the fifth dimension – like a domain-specific one. We could also observe that as we increase the number of factors in the model, the correlation of the first factor with the ISEI becomes lower and lower.

*Wikinews*

To test the robustness of our results we repeated our analysis on a different corpus, namely on the Wikinews corpus. In this corpus, the most frequent occupation was the editor, but judge, politician, or lawyer were also frequent, such as journalist, writer and singer. Most of these are higher prestige occupations, which are related to the domains of politics, media and culture. For comparison, we run the same factor analyses as on the CC-based embedding. The results were more similar than we expected. In the case of the 3-factor solution, the correlations of the first factor scores of the two corpora was 0.97, the correlation of the second factors was 0.93 and of the third factors it was 0.82. These results suggest that the factors in the two corpora show similar structure of the occupations.

With a more detailed analysis we could find minor differences between the first factors of the Wikinews and CC corpus. Some manual-labor occupations like locksmith and dishwasher got higher scores in the Wikinews corpus and some of the literature and art related occupations, like poet, novelist, composer or painter scored higher in the CC corpus. Table 4 presents the occupations with the highest loadings in each dimension.

| First factor | | | | Second factor | | | | Third factor | | | |
|---|---|---|---|---|---|---|---|---|---|---|---|
| Highest loadings | | Lowest loadings | | Highest loadings | | Lowest loadings | | Highest loadings | | Lowest loadings | |
| rank order | ISEI | rank order | ISEI | rank order | ISEI | rank order | ISEI | rank order | ISEI | rank order | ISEI |
| chairperson | 71.29 | bartender | 25.04 | biologist | 80.46 | bender | 25.78 | secretary | 44.94 | acrobat | 37.59 |
| chancellor | 70.34 | dressmaker | 23.47 | mathematician | 81.78 | dishwasher | 16.5 | prosecutor | 86.72 | assembler | 27.91 |
| dean | 65.01 | shopkeeper | 35.34 | zoologist | 80.46 | maid | 14.21 | mayor | 68.77 | bricklayer | 22.57 |
| advocate | 86.72 | blacksmith | 25.63 | philosopher | 83.81 | barman | 25.04 | governor | 68.77 | jeweller | 28.12 |
| commentator | 72.83 | hairdresser | 31.08 | physicist | 84.61 | barista | 25.04 | chairperson | 71.29 | goldsmith | 28.12 |
| ecologist | 80.46 | roofer | 22.16 | botanist | 80.46 | courier | 30.34 | commissioner | 78.76 | shoemaker | 18.07 |
| director - general | 71.29 | barmaid | 25.04 | historian | 83.81 | waiter | 25.04 | senator | 68.77 | potter | 24.43 |
| professor | 85.41 | locksmith | 33.16 | ornithologist | 80.46 | janitor | 21.82 | attorney | 86.72 | beekeeper | 28.04 |
| historian | 83.81 | carpenter | 26.62 | geographer | 83.09 | clerk | 43.33 | treasurer | 73.38 | optician | 59.85 |
| sociologist | 83.09 | barman | 25.04 | ecologist | 80.46 | stewardess | 46.76 | lawyer | 86.72 | roofer | 22.16 |
| biographer | 72.83 | bricklayer | 22.57 | sociologist | 83.09 | babysitter | 24.98 | chancellor | 70.34 | tanner | 28.08 |
| editor | 72.83 | waiter | 25.04 | writer | 72.83 | barmaid | 25.04 | dean | 65.01 | tattooist | 50.15 |
| governor | 68.77 | waitress | 25.04 | geologist | 86.81 | waitress | 25.04 | ambassador | 78.76 | weaver | 28.95 |
| geographer | 83.09 | plumber | 29.16 | poet | 72.83 | cleaner | 16.38 | councillor | 68.77 | welder | 28.52 |
| marshal | 60.92 | plasterer | 18.02 | novelist | 72.83 | receptionist | 39.02 | professor | 85.41 | plasterer | 18.02 |

Table 4 Occupations with highest and lowest loadings, 3-factor solution, Wikinews

The interpretation of the first three dimensions is quite similar to the ones in the CC corpus. The first factor shows a mixed organizing pattern built of power and knowledge. In the case of the second factor, the science related occupations scores high. The dimension behind the third factor is about power level and organizational capacity. The correlation of the first dimension with the ISEI score was 0.71 (see Figure A1) and we found that with the above described methodology, 74 percent of the occupation pairs are in the same order in the Wikinews based first factor hierarchy and on the ISEI scale. In addition to the similarities, we also found differences: some animal- and farm related occupations (e.g., breeder, fisher, planter) score much higher on the semantic scale, and some health-related occupations (e.g., doctor, surgeon, dentist, pharmacist) score higher on the ISEI scale.

We have also tested the 4- and 5-factor solutions here. Similar to the result of the CC corpus, the 4th factor can be interpreted as the gender dimension: occupations like nanny, hairdresser, receptionist, babysitter or waitress score high there. Just as in the CC corpus, the 5th dimension was a domain related one. It is interesting, however, that in the current (Wikinews) corpus, it was not the health domain, which characterized the scale, but the domain of media and culture, with highly scored occupations like novelist, poet, singer, composer, dramatist, lyricist, or writer.

*Wikinews with sub-word information*

The last word embedding we tested was also built on the Wikinews corpus, but the training phase of this model also took into account sub-word information. With this solution, partly identical words or words with the same root are closer to each other in the vector space. The same 3-factor solution was applied here and the interpretation of the three factors is on the whole the same as in the previous cases. (For more details about these factors, see Table A3 in the Appendix).

The interpretation of the factors showed that institutional power is an important aspect in the first factor, but knowledge also matters there. The second factor was related to the knowledge and educational level associated with the occupations, while the third factor was scaled on the power levels and organizational capacities of the occupations. This later factor is close to the domain of politics.

The correlation of the first dimension with the ISEI score was 0.78. According to the rank order, 77 percent of the occupation pairs were the same on both scales, namely in the first factor of this corpus and the ISEI. The occupations, which are much higher on the semantic scale are rancher, planter, and astrologist. Other occupations are underestimated compared to the ISEI: such as in the case of the CC corpus, these are domain specific occupations. Some of them are health-related occupations, such as dentist, doctor, pharmacist and surgeon; some are financial occupations, like banker or accountant; and some are judicial system related occupations like judge, lawyer or solicitor.

We also tested the 4- and 5-factor solution here. The 4th factor showed the gender dimension again with high scores at occupations like nanny, hairdresser, receptionist, babysitter and waitress. The 5th factor was again a domain-related one, namely the domain of media and culture with high scores at occupations like novelist, poet, singer, composer, dramatist, lyricist, and writer – just like in the case of the Wikinews corpus.

*Robustness – Stability of occupational positions in different vector spaces*

The correlation of the factor loadings across different embeddings seems to be really strong. The correlations of the first factor scores of the CC and Wikinews embeddings was 0.97, between the second factors it was 0.93, and between the third factors 0.82. These results provide strong evidence for the robustness of the results and implicate that occupation positions are quite stable across different corpora.

To further test this stability, we wanted to compare the positions of the occupations in the different vector spaces. In order to do this, we had to align the Wikinews vector space to the CC vector space with Procrustes rotation the way we described earlier. As we stated above, in this aligned vector space, the cosine similarities of the words are the same as before the alignment, but we are able to calculate the similarity of the same occupation between the two vector spaces. The average similarity of the occupations between the two corpora was 0.79. There is no clear threshold what similarity level can be interpreted as 'strong', but we can observe that, only close concepts have a similarity value around 0.7. An intuitive example for this in the CC embedding is dog breeds, like Labrador and Beagle, which have the similarity value around 0.7. Although the average similarity measure implicates high stability between the embeddings, there are some occupations where we found lower – but in absolute values still high – similarities. Occupations with the lowest similarity scores (between 0.65 and 0.7) were the following: masseur, dishwasher, rheumatologist, manicurist, zookeeper, editor, bender, locksmith, dentist, and tanner. We cannot observe a clear organizing principle, but some of these occupations are quite rare now, like the tanner or the bender.

We calculated the correlation of Wikinews frequency of occupations and the stability measure, and its value was 0.59. This result is parallel with earlier findings, namely that those words are stable across time, which are frequent (Hamilton – Leskovec – Jurafsky 2016). Our results show that it is not only applicable for temporal analysis, but also for the analysis of different corpora (and embeddings) created approximately at the same time. Stability also positively correlated with the ISEI score (Pearson $r=0.36$, $p=0.00$). The direction of the correlation suggests that the positions of more prestigious occupations are more stable across corpora, but this result should be treated with caution, as this effect partly exist because more prestigious occupations are also more frequent (at least in the two corpora we used). However, even after controlling for the frequencies of the words, the correlation still remains significant ($r=0.19$, $p=0.000$) between ISEI score and stability.

# Discussion

We raised two questions about the usefulness of word embedding based semantic analysis related to the description of occupational structure in particular occupational rankings. Are the results comparable with standard results and is it possible to gain additional insights about the social positions of occupations? Both questions raised at the beginning of the paper have been given affirmative answers. The results show fundamental similarity of the social structure obtained from text analysis to the structure described by Ganzeboom and Treiman (1996). But a more detailed analysis also reveals some differences.

Our paper focused more on methodological aspects and we put less emphasis on the substantive analysis of the results. But the first – superficial – analysis revealed an interesting dimension of the occupation structure: the power and organizational aspect. As far as we know the importance of this factor is not discussed in the main line of stratification literature in sociology. It has been widely discussed (Johnson 2016) that power is an important component of the prestige of an occupation. But our results indicate the interplay between knowledge and organizational capacity. In the 3-factor solution, each is characterized by the presence of one or both of these, and power presents itself as a combination of knowledge and organizational capacity. It is not a surprise that knowledge, also in itself, is a fundamental dimension, but it does seem quite novel, that organizational capacity, also in itself, is a contributing dimension. Freidson (1984) distinguishes two types of elites: knowledge and administrative elites in his classic work. Waring (2014) re-apprised Freidson model and added two extra elite types, corporate and governance elite. Our third factor mirror the importance of this governance elite as an important factor that structure the occupational space.

The results proved quite stable, as repeating the analyses on two different corpora yielded strongly similar results. Correlations of the factors between the two corpora were high and substantively significant. After the alignment of the second corpus on the first one, we found strong similarities in the positions of the occupations across corpora. Although we don't have data for measuring other stability indicators, but we know from other studies (Hamilton – Leskovec – Jurafsky 2016) that concept stability is lower for words, which are frequently used in different environments – that is called polysemy in linguistic. It is also known that the position of a concept changes over time (Kozlowski et al. 2019), so further analysis may also take into account the time period during which the original corpora were collected.

We decided to use pre-trained corpora in this paper and not trained unique word embeddings. These pre-trained corpora are available for everyone, so it is pretty easy to reproduce our results and make further steps in this area. One shortcoming of this approach that we could not narrow the geographical focus of the results, and we could not influence what type of texts are included in the training set. However, previous studies showed (Treimann 1977) that prestige scores are highly correlated in developed countries. So our general approach might not lead to significant biases. The fact also confirms the validity of the results, that results coming from different corpora and word embedding was similar.

Nevertheless, it could be a logical step to repeat this analysis with self-trained word-embeddings, where we have stronger control of the selected texts. Training our models has a further advantage; we could pre-process the texts before calculating the vector spaces. For social science analysis, pre-processed texts could work better as the information is focused here, and there is less noise in those texts. We could also add bi-grams to the model, which might be essential to catching the two-word length occupations like "social scientist." Further studies are needed to understand how pre-preprocessing influences word embedding features and how this affects any social science-related analysis.

Our paper presents exploratory research using textual data, with fairly new methods in the social sciences although it has already been demonstrated that unsupervised learning methods such as the analysis of word embeddings are able to find interesting patterns and generate new hypothesis (Nelson 2020). Both qualitative and quantitative approaches are needed to fully exploit this potential in understanding societies.

**List of abbreviations**

NLP: Natural Language Processing
ISEI: International Socio-Economic Index
SIOPS: Standard International Occupational Prestige Scale
CC: Common Crawl
RMSR: Average residuals for the similarity matrix
OSPS: Occupation Semantic Position Scale

**Declarations**

**Availability of data and material:** The pre-trained word vectors are available here:

Common Crawl: http://commoncrawl.org

Wikinews: https://fasttext.cc/docs/en/english-vectors.html

**Competing interests**

None of the authors have any competing interests in the manuscript.

**Funding**

The work of Zoltan Kmetty was funded by the Premium Postdoctoral Grant of the Hungarian Academy of Sciences.

The work of Julia Koltai was funded by the Premium Postdoctoral Grant of the Hungarian Academy of Sciences.

**Authors' contributions**

    Zoltán Kmetty: Concept, computations, analysis, discussion

    Júlia Koltai: Concept, theoretical Part, discussion

    Tamás Rudas: Concept, theoretical Part, discussion

**Acknowledgements**
29

**References**

Bales, R. F. (1950). A set of categories for the analysis of small group interaction. American Sociological Review 15(2), 257–63.

Bolukbasi, T., Chang, K. W., Zou, J. Y., Saligrama, V., & Kalai, A. T. (2016). Man is to computer programmer as woman is to homemaker? debiasing word embeddings. In Advances in neural information processing systems (pp. 4349-4357).

Bukodi, E., Dex, S., & Goldthorpe, J. H. (2011). The conceptualisation and measurement of occupational hierarchies: a review, a proposal and some illustrative analyses. Quality & Quantity, 45(3), 623-639.

Caliskan, A., Bryson, J. J., & Narayanan, A. (2017). Semantics derived automatically from language corpora contain human-like biases. Science, 356(6334), 183-186.

Chan, T. W., & Goldthorpe, J. H. (2004). Is there a status order in contemporary British society? Evidence from the occupational structure of friendship. European Sociological Review, 20(5), 383-401.

Connelly, R., Gayle, V., & Lambert, P. S. (2016). A Review of occupation-based social classifications for social survey research. Methodological Innovations, 9, 2059799116638003.

Devlin, J., Chang, M. W., Lee, K., & Toutanova, K. (2018). Bert: Pre-training of deep bidirectional transformers for language understanding. *arXiv preprint arXiv:1810.04805*.

Erikson, R., Goldthorpe, J. H., & Portocarero, L. (1979). Intergenerational class mobility in three Western European societies: England, France and Sweden. The British Journal of Sociology, 30(4), 415-441.

Evans, J. A., & Aceves, P. (2016). Machine translation: mining text for social theory. Annual Review of Sociology, 42, 21-50.

Firth, J. R. (1957). A synopsis of linguistic theory. Studies in linguis-tic analysis. Oxford: Blackwell.

Freidson, E. (1984). The changing nature of professional control. *Annual review of sociology*, *10*(1), 1-20.

Ganzeboom, H. B., & Treiman, D. J. (1996). Internationally comparable measures of occupational status for the 1988 International Standard Classification of Occupations. Social science research, 25(3), 201-239.

Garg, N., Schiebinger, L., Jurafsky, D., & Zou, J. (2018). Word embeddings quantify 100 years of gender and ethnic stereotypes. Proceedings of the National Academy of Sciences, 115(16), E3635-E3644.

Goldthorpe, J. H., & Hope, K. (1972). Occupational grading and occupational prestige. Social Science Information, 11(5), 17-73.

Goldthorpe, J. H., Halsey, A. H., Heath, A. F., Ridge, J. M., Bloom, L., & Jones, F. L. (1982). Social mobility and class structure in modern Britain.

Hamilton, W. L., Leskovec, J., & Jurafsky, D. (2016). Diachronic word embeddings reveal statistical laws of semantic change. arXiv preprint arXiv:1605.09096.

Hays, D.C. (1960): Automatic Content Analysis. Santa Monica, CA, Rand Corp.

Johnson, T. J. (2016). *Professions and Power (Routledge Revivals)*. Routledge.

Joseph, K., & Morgan, J. H. (2020). When do Word Embeddings Accurately Reflect Surveys on our Beliefs About People?. *arXiv preprint arXiv:2004.12043*.

Joulin, A., Grave, E., Bojanowski, P., & Mikolov, T. (2016). Bag of tricks for efficient text classification. arXiv preprint arXiv:1607.01759.

Kozlowski, A. C., Taddy, M., & Evans, J. A. (2019). The Geometry of Culture: Analyzing the Meanings of Class through Word Embeddings. American Sociological Review, 84(5), 905-949.

Lazer, D. & Radford, J. (2017): Data ex Machina: Introduction to Big Data. Annual Review of Sociology, 43(1):19–39.

Meraviglia, C., Ganzeboom, H. B., & De Luca, D. (2016). A new international measure of social stratification. *Contemporary Social Science*, *11*(2-3), 125-153.

Mikolov, T., Chen, K., Corrado, G., & Dean, J. (2013). Efficient estimation of word representations in vector space. arXiv preprint arXiv:1301.3781.

Nelson, L. K. (2020). Computational grounded theory: A methodological framework. Sociological Methods & Research, 49(1), 3-42.

Németh, R. & Koltai, J. (2019) Discovering sociological knowledge through automated text analytics In: Rudas, Tamás – Péli, Gábor (eds.) Pathways Between Social Science and Computational Social Science – Therories, Methods and Interpretations. New York, NY, Springer. (forthcoming)

Pennington, J., Socher R., and Manning C. D.(2014). GloVe: Global Vectors for Word Representation

Prandy, K., & Lambert, P. (2003). Marriage, social distance and the social space: an alternative derivation and validation of the Cambridge Scale. Sociology, 37(3), 397-411.

Revelle, W. (2018) psych: Procedures for Personality and Psychological Research, Northwestern University, Evanston, Illinois, USA, https://CRAN.R-project.org/package=psych Version = 1.8.12.

Rose, D., & Harrison, E. (2007). The European socio-economic classification: a new social class schema for comparative European research. European Societies, 9(3), 459-490.

Savage, M., & Burrows, R. (2007). The Coming Crisis of Empirical Sociology. Sociology: A Journal of the British Sociological Association, 41, 885–899.

Schakel, A. M., & Wilson, B. J. (2015). Measuring word significance using distributed representations of words. arXiv preprint arXiv:1508.02297.

Treiman, D.J. (1977). Occupational Prestige in Comparative Perspective. Academic Press, New York

Waring, J. (2014). Restratification, hybridity and professional elites: questions of power, identity and relational contingency at the points of 'professional–organisational intersection'. *Sociology Compass*, *8*(6), 688-704.

# Appendix

accompanist, accountant, acrobat, actor, actuary, admiral, advocate, agriculturist, agrologist, agronomist, allergist, ambassador, anaesthesiologist, anatomist, animator, appraiser, archaeologist, architect, assembler, astrologer, astronaut, athlete, attorney, auditor, babysitter, baker, ballerina, banker, barber, barista, barkeeper, barmaid, barman, bartender, beekeeper, bender, biographer, biologist, bishop, blacksmith, blocklayer, boatman, bodyguard, bookkeeper, bookmaker, botanist, boxer, brazier, breeder, brewer, bricklayer, broker, butcher, cardiologist, carer, carpenter, cellist, ceo, chairperson, chancellor, chaplain, chef, chemist, cleaner, clerk, coalman, coastguard, coder, comedian, commentator, commissioner, composer, congressman, congresswoman, constable, cook, copywriter, coroner, corporal, councillor, courier, curator, dancer, dean, dentist, director-general, dishwasher, dockmaster, doctor, doorkeeper, dramatist, dressmaker, driller, driver, dustman, ecologist, editor, electrician, environmentalist, etcher, farmer, firefighter, fireman, fisher, flamecutter, footballer, forger, friar, furrier, gaoler, gardener, geodesist, geographer, geologist, goatherd, goldsmith, governor, grazier, grocer, hairdresser, head-teacher, historian, hooker, providing sexual services, housemaid, innkeeper, janitor, jeweller, journalist, judge, juggler, lawyer, lecturer, librarian, locksmith, lyricist, macroeconomist, maid, managing-director, manicurist, marketer, marshal, masseur, mathematician, mayor, mechanic, meteorologist, midwife, miner, money-lender, monk, nanny, neurologist, nightwatchman, novelist, nurse, optician, ornithologist, painter, paratrooper, parliamentarian, pastry-cook, pharmacist, philosopher, photographer, physicist, physiotherapist, planter, plasterer, plumber, poet, policeman, policewoman, politician, postman, postmaster, potter, priest, professor, programmer, proofreader, prosecutor, prostitute, psychiatrist, psychologist, psychotherapist, publicist, rabbi, radiographer, rancher, receptionist, rector, retailer, rheumatologist, roofer, sailor, secretary, senator, setter-operator, shepherd, shoe-polisher, shoemaker, shopkeeper, signwriter, singer, sociologist, soldier, solicitor, sommelier, sous-chef, stationmaster, statistician, steward, stewardess, stonecutter, storekeeper, surgeon, tailor, tanner, tattooist, telemarketer, telephonist, tiler, translator, treasurer, typist, vendor, waiter, waitress, weaver, webmaster, welder, writer, zookeeper, zoologist

Table A1. List of occupations

| Factor 1 | Factor 2 |
|---|---|
| curator | historian |
| editor | biologist |
| geographer | zoologist |
| professor | sociologist |
| sociologist | geographer |
| biologist | physicist |
| chairperson | journalist |
| historian | ornithologist |
| environmentalist | lecturer |
| commentator | writer |

Table A2. Occupations with highest loadings, 2-factor solution, CC

| Factor 1 | Factor 2 | Factor 3 |
|---|---|---|
| professor | biologist | commissioner |
| congresswoman | zoologist | secretary |
| biographer | ecologist | mayor |
| CEO | physicist | chancellor |
| ecologist | ornithologist | chairperson |
| neurologist | sociologist | prosecutor |
| director-general | mathematician | governor |
| chairperson | geographer | senator |
| chancellor | botanist | attorney |
| dean | geologist | treasurer |

Table A3. Occupations with highest loadings, 3-factor solution, Wikinews_subwords

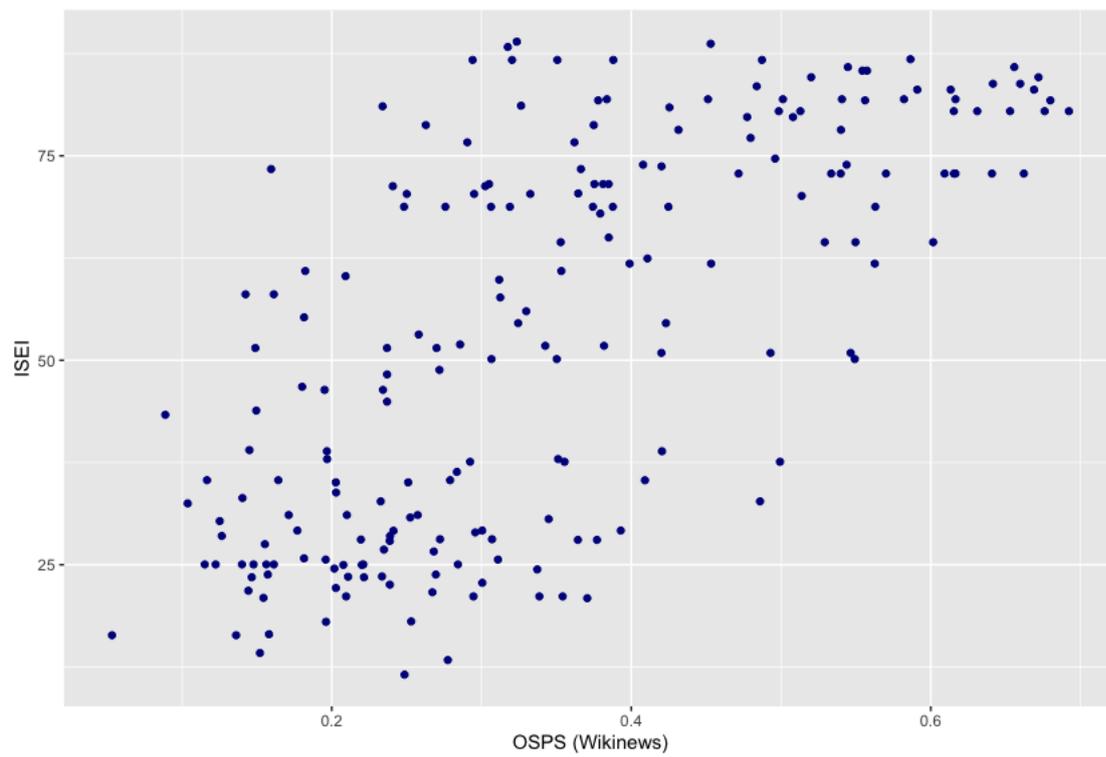

Figure A1 Scatterplot of word embedding based occupation prestige score (Occupation Semantic Position Scale – OSPS) from the CC vector space and ISEI

ENDNOTES

---

i http://commoncrawl.org
ii https://fasttext.cc/docs/en/english-vectors.html